# AVP-Pro: An Adaptive Multi-Modal Fusion and Contrastive Learning Approach for Comprehensive Two-Stage Antiviral Peptide Identification

Xinru Wen[1], Weizhong Lin[1], zi liu[1] and Xuan Xiao[1,2*]

1. School of Information Engineering, Jingdezhen Ceramic University, Jingdezhen Jiangxi, China.

2. School of Information Engineering, Jingxi Art & Ceramics Technology Institute, Jingdezhen Jiangxi, China.

Corresponding author: Xuan Xiao: jdzxiaoxuan@163.com , xiaoxuan@jci.edu.cn

# Abstract

The accurate identification of antiviral peptides (AVPs) is crucial for novel drug development. However, existing methods still have limitations in capturing complex sequence dependencies and distinguishing confusing samples with high similarity. To address these challenges, we propose AVP-Pro, a novel two-stage predictive framework that integrates adaptive feature fusion and contrastive learning. To comprehensively capture the physicochemical properties and deep-seated patterns of peptide sequences, we constructed a panoramic feature space encompassing 10 distinct descriptors and designed a hierarchical fusion architecture. This architecture integrates self-attention and adaptive gating mechanisms to dynamically modulate the weights of local motifs extracted by CNNs and global dependencies captured by BiLSTMs based on sequence context. Targeting the blurred decision boundary caused by the high similarity between positive and negative sample sequences, we adopted an Online Hard Example Mining (OHEM)-driven contrastive learning strategy enhanced by BLOSUM62. This approach significantly sharpened the model's discriminative power. Model evaluation results show that in the first stage of general AVP identification, the model achieved an accuracy of 0.9531 and an MCC of 0.9064, outperforming existing state-of-the-art (SOTA) methods. In the second stage of functional subtype prediction, combined with a transfer learning strategy, the model realized accurate classification of 6 viral families and 8 specific viruses under small-sample conditions. AVP-Pro provides a powerful and interpretable new tool for the high-throughput screening of antiviral drugs. To further enhance accessibility for users, we have developed a user-friendly web interface, which is available at https://wwwy1031-avp-pro.hf.space.

# Introduction

Given that the escalating drug resistance crisis associated with viral diseases has emerged as one of the most pressing threats to global public health, there is an urgent imperative for the development and implementation of novel therapeutic modalities[1-3]. In contrast to traditional small-molecule agents, antiviral peptides (AVPs) exert their antiviral effects via distinct mechanisms of action, including the inhibition of viral attachment to host cell surfaces and the disruption of key viral replication pathways. Endowed with high specificity, low inherent toxicity, and a diminished propensity to elicit drug resistance, AVPs have garnered considerable attention as highly promising alternatives to conventional antiviral therapeutics[4-6]. The successful clinical translation of peptide-based drugs, such as enfuvirtide (an HIV fusion inhibitor) and boceprevir (a therapeutic agent for hepatitis C), further corroborates their substantial therapeutic potential in clinical settings[7, 8].

Beyond their high specificity and low toxicity, the clinical utility of antiviral peptides (AVPs) is largely contingent upon their precision in targeting specific viral pathogens. Distinct AVPs typically inhibit the lifecycle of specific viruses via unique mechanisms. For instance, certain peptides can specifically abrogate the binding of hepatitis C virus (HCV) to host hepatocytes, while others focus on suppressing the replication of human immunodeficiency virus (HIV) or interfering with the infection process of herpes simplex virus (HSV). Consequently, to enhance the practical relevance of predictive models, they must not only determine whether a peptide exhibits antiviral activity but also explicitly delineate the specific class of viruses it targets,such as coronaviruses and SARS-CoV-2[9].

However, the experimental identification of AVPs is both time-consuming and labor-intensive[10]. To address this gap, machine learning (ML) has been extensively utilized to expedite the screening of potential AVP candidates from large-scale peptide libraries. Early studies primarily relied on manually extracted features combined with conventional ML algorithms for predictive modeling. For instance, Thakur et al. pioneered AVPpred[11], which employs support vector machines (SVMs) as classifiers and utilizes amino acid composition (AAC) and physicochemical properties as input features for AVP prediction. Subsequently, Chang et al. [12]improved predictive performance by adopting random forests (RFs) based on aggregation propensity features, while Lissabet et al. introduced AntiVPP 1.0[13]—a model that integrates global features such as electrostatic charge and hydrophobicity. To further enhance model robustness, ensemble learning strategies have emerged as a viable approach. Among these, Schaduangrat et al. developed Meta-iAVP[14], a meta-predictor that aggregates six distinct learning hypotheses. Additionally, Akbar et al. [15]constructed an optimized ensemble classifier by integrating discrete wavelet transform with SHAP-based feature selection.

Despite these advancements, conventional ML-based predictors continue to encounter substantial limitations. First, they exhibit a strong reliance on manual feature engineering (e.g., AAC, PseAAC). This process not only demands profound domain expertise but also frequently fails to capture deep evolutionary insights or complex nonlinear sequence dependencies. Second, most early studies were confined to binary classification (AVP vs. non-AVP), overlooking the critical need to identify activity specific to particular viral families—an essential requirement for precision medicine. Furthermore, these models rely on outdated benchmark datasets (e.g., those compiled in 2012) that fail to capture the diversity and complexity of sequences in modern AVP databases. This limitation restricts their applicability to newly emerging viral strains.

To address these bottlenecks, deep learning (DL) has been introduced into the field, with the aim of automatically learning high-level representations from raw sequences for predictive modeling. Li et al. proposed DeepAVP[16], a dual-channel network integrating CNNs [17]and LSTMs[18] to capture local motifs and global patterns. In contrast, ENNAVIA [19] leverages neural networks for physicochemical feature learning. Recently, two-stage frameworks such as FFMAVP [20] and AVPIden [21]have been developed to concurrently address general identification and subclass prediction tasks. Furthermore, AVP-IFT [22] and AVP-HNCL [23] represent the state-of-the-art advances in this field achieved via the application of deep representation learning. Nevertheless, existing DL methods still exhibit notable limitations: (1) Feature fusion is often static. most models (e.g., FFMAVP) rely on simple feature concatenation, overlooking the dynamic contextual differences in the importance of local versus global signals across distinct sequences. (2) Weak decision boundaries. conventional approaches typically employ random negative sampling or standard loss functions, which fail to effectively discriminate "hard" negative samples that are compositionally similar to AVPs, resulting in elevated false positive rates. (3) Scarce subclass data. although some models attempt subclass prediction, they often struggle with extreme data imbalance and scarcity within specific viral families, leading to compromised generalization capabilities.

To overcome these challenges, we propose AVP-Pro, a novel two-stage deep learning framework. AVP-Pro synergistically integrates three core innovations to achieve robust AVP identification and accurate subclass prediction.

(1) Adaptive multimodal fusion. We designed a hierarchical fusion architecture. This architecture leverages self-attention mechanisms [24, 25] to reinforce local motifs extracted by CNNs and global dependency features captured by BiLSTMs. Subsequently, an Adaptive Gating Mechanism [26] dynamically modulates the weights of these two components based on sequence context, thereby outperforming traditional static concatenation approaches.

(2) Biology-guided data augmentation. We abandon random noise and adopt a "suboptimal" mutation strategy guided by the BLOSUM62 substitution matrix [27, 28]. BLOSUM62 encapsulates evolutionary probability information of amino acid substitutions, ensuring that the generated samples maintain biological plausibility while enriching data diversity.

(3) Prototype-based OHEM strategy. To enhance the model's ability to discriminate between highly similar confounding samples, we implement a contrastive learning strategy driven by Online Hard Example Mining (OHEM)[29]. This approach actively identifies and learns from ambiguous samples that are difficult to classify, substantially improving the model's specificity.

In summary, AVP-Pro not only enables efficient AVP identification but also leverages transfer learning [30, 31] to accurately predict their functional subclasses, providing a powerful and interpretable tool for the rational screening of antiviral peptides.

# RESULTS AND DISCUSSION

## Model performance

In this study, to rigorously validate the model's generalization performance, we strictly partitioned all experimental datasets into training and independent test sets at a 4:1 ratio. Model parameters were optimized on the training set, while all subsequent performance evaluation results were derived exclusively from the independent test set, thereby ensuring the objectivity and fairness of the assessment.

In the first stage of the study, the performance of AVP-Pro on two benchmark datasets was comprehensively evaluated (see Table 1 for details). The model exhibited exceptional discriminative capability on the independent test sets of both Set 1-nonAVP and Set 2-nonAMP. Specifically, on the highly challenging Set 1-nonAVP dataset, our model achieved accuracy (ACC) of 0.9531, sensitivity (SN) of 0.9644, specificity (SP) of 0.9418, and a Matthews Correlation Coefficient (MCC) of 0.9064. On the Set 2-nonAMP dataset, AVP-Pro delivered perfect predictions, with all key metrics (ACC, SN, SP, MCC) reaching 1.0000. These results fully demonstrate that the model can robustly distinguish AVPs from non-AVPs regardless of the type of negative samples, highlighting its strong generalization ability.

In the second phase, using the same independent test set partitioned at a 4:1 ratio, we extended the prediction task to more granular functional subtype identification. This task encompasses six major viral families, including Coronaviridae, Retroviridae, Herpesviridae, Paramyxoviridae, Orthomyxoviridae, and Flaviviridae, as well as eight specific viruses, including Feline Immunodeficiency Virus (FIV), Human Immunodeficiency Virus (HIV), Hepatitis C Virus (HCV), Human Parainfluenza Virus Type 3 (HPIV3), Herpes Simplex Virus Type 1 (HSV1), Influenza A Virus (INFVA), Respiratory Syncytial Virus (RSV), and Severe Acute Respiratory Syndrome Coronavirus (SARS-CoV). Given the prevalent sample imbalance in these sub-datasets, we constructed a multi-dimensional evaluation framework to avoid biases from a single metric: G-mean was introduced to objectively assess the performance balance between majority and minority classes. AUROC was incorporated to evaluate the model's comprehensive discriminative power across different classification thresholds. AUPRC was specifically adopted as it better reflects the model's precise retrieval ability for minority classes (i.e., specific antiviral activities) when positive samples are scarce.

Table 2 presents the detailed performance on viral family classification. Benefiting from the optimization of the imbalance learning strategy, the model performed excellently across all six viral families, with accuracies ranging from 0.8872 to 0.9887. Notably, the model achieved perfect specificity (1.0000) for both Orthomyxoviridae and Paramyxoviridae, with G-mean scores of 0.8367 and 0.9478, respectively. The classification performance for the eight specific target viruses is shown in Table 3. Our model also demonstrated robust performance in these tasks, with accuracies distributed between 0.9041 and 1.0000, and G-mean scores fluctuating within the high range of 0.8157 to 1.0000. Among these, the prediction results for HPIV3 were particularly outstanding, with all evaluation metrics reaching a perfect score of 1.0000, demonstrating the model's outstanding performance on this specific task. Additionally, for key viruses such as HCV, RSV, and SARS-CoV, the model exhibited high overall robustness, with AUROC scores of 0.9642, 0.9517, and 0.9652, respectively, and AUPRC scores maintaining high levels of 0.9442, 0.8669, and 0.9227. These data strongly confirm the effectiveness and reliability of AVP-Pro in accurately identifying peptides targeting specific viruses.

**Table1:** Performance summary on independent dataset of non-AVP and non-AMP datasets

| Dataset | ACC | MCC | SN | SP |
|---|---|---|---|---|
| set 1-nonAVP | 0.9531 | 0.9064 | 0.9644 | 0.9418 |
| set 2-nonAMP | 1.0000 | 1.0000 | 1.0000 | 1.0000 |

**Table2:** Performance summary on independent datasets of viral family datasets.

| Viral family | Accuracy | MCC | Sensitivity | Specificity | G-mean | AUROC | AUPRC |
|---|---|---|---|---|---|---|---|
| Coronaviridae | 0.9850 | 0.8838 | 0.8919 | 0.9919 | 0.9406 | 0.9619 | 0.9011 |
| Flaviviridae | 0.9417 | 0.7992 | 0.7071 | 0.9954 | 0.8389 | 0.9334 | 0.8771 |
| Herpesviridae | 0.9455 | 0.6867 | 0.5690 | 0.9916 | 0.7511 | 0.9063 | 0.7602 |
| Orthomyxoviridae | 0.9887 | 0.8318 | 0.7000 | 1.0000 | 0.8367 | 0.9696 | 0.8262 |
| Paramyxoviridae | 0.9887 | 0.9418 | 0.8983 | 1.0000 | 0.9478 | 0.9870 | 0.9650 |
| Retroviridae | 0.8872 | 0.7650 | 0.8194 | 0.9335 | 0.8746 | 0.9356 | 0.9215 |

**Table3:** Performance summary on independent datasets of targeted virus datasets.

| Targeted virus | Accuracy | MCC | Sensitivity | Specificity | G-mean | AUROC | AUPRC |
|---|---|---|---|---|---|---|---|
| HCV | 0.9662 | 0.8838 | 0.9817 | 0.8947 | 0.9372 | 0.9642 | 0.9442 |
| HIV | 0.9041 | 0.7876 | 0.9741 | 0.7730 | 0.8677 | 0.9341 | 0.8960 |
| HPIV3 | 1.0000 | 1.0000 | 1.0000 | 1.0000 | 1.0000 | 1.0000 | 1.0000 |
| HSV1 | 0.9624 | 0.7368 | 0.9836 | 0.7209 | 0.8421 | 0.9454 | 0.8183 |
| INFVA | 0.9906 | 0.8631 | 0.9980 | 0.8000 | 0.9183 | 0.9837 | 0.8970 |
| RSV | 0.9906 | 0.8757 | 0.9980 | 0.8182 | 0.9036 | 0.9517 | 0.8669 |
| SARS-CoV | 0.9907 | 0.9023 | 0.9982 | 0.8571 | 0.9249 | 0.9652 | 0.9227 |
| FIV | 0.9868 | 0.7784 | 0.6667 | 0.9981 | 0.8157 | 0.9474 | 0.8229 |

## Compare with other existing AVP prediction tools

To further evaluate the performance of AVP-Pro, we conducted comprehensive performance benchmarking against state-of-the-art predictive tools. First, we performed a head-to-head comparison on the benchmark independent datasets shared by AVP-HNCL and AVP-IFT. As presented in Table 4, AVP-Pro achieved comprehensive superiority across all key evaluation metrics. Particularly on the highly challenging Set 1-nonAVP dataset—where sample discrimination poses extreme difficulty—our model achieved accuracy (ACC) gains of 1.8% and 3.1% over AVP-HNCL and AVP-IFT, respectively. For the MCC metric, which provides a more robust reflection of overall discriminative capability, the performance improvements reached 3.8% and 6.8% relative to the same two models. These significant performance gains strongly validate the superiority of our hierarchical attention fusion architecture and OHEM strategy in handling complex samples. Furthermore, on the Set 2-nonAMP dataset, AVP-Pro achieved theoretically perfect predictive performance (with both ACC and MCC scores at 1.0000), demonstrating notable specificity.

Beyond its outstanding performance on the aforementioned standard benchmark datasets, we further validated AVP-Pro's generalization capability and robustness across diverse data distributions by comparing it with two other leading predictive tools, including FFMAVP and AVPIden. For FFMAVP, We adopted the benchmark data partitioning scheme provided in its official codebase. This partitioning corresponds to the Best Fold within its 4-fold cross-validation (4-fold CV), featuring distinct training and test sets. To ensure rigorous comparison, we strictly adhered to this official partitioning scheme: we trained AVP-Pro using the provided training data and conducted the final validation on the specified test data. And for AVPIden, we adopted its consistent 4-fold cross-validation strategy. As presented in Table 5, AVP-Pro demonstrated remarkable superiority once again. On FFMAVP's test set, our model achieved an accuracy (ACC) of 0.9689 and a MCC of 0.9099, outperforming FFMAVP by 2.5% and 7.5%, respectively. Particularly in terms of the sensitivity (SN) metric, AVP-Pro reached a score of 0.9522, which was significantly higher than FFMAVP's 0.8802, indicating our model's superior ability to capture latent antiviral features.In the cross-validation testing for AVPIden, AVP-Pro also exhibited exceptional performance, achieving an ACC of 0.9492, which was markedly superior to AVPIden's 0.9116. Notably, thanks to the mining of hard negative samples enabled by the OHEM strategy, our model achieved a substantial improvement in specificity (SP) (0.9707 vs. 0.9044), thus significantly reducing the risk of false positives. This consistent, superior performance across datasets fully demonstrates that AVP-Pro is not only effective on specific benchmarks but also possesses strong generalization potential for addressing diverse data distributions.

**Table4.** Comparison of the Performance of Existing Methods on Set 1-nonAVP and Set 2-nonAMP.

| Dataset | method | ACC | SN | SP | MCC |
|---|---|---|---|---|---|
| set 1-nonAVP | AVP-HNCL | 0.9362 | 0.9550 | 0.9174 | 0.8730 |
| | AVP-IFT | 0.9240 | 0.9343 | 0.9137 | 0.8482 |
| | **our method** | **0.9531** | **0.9644** | **0.9418** | **0.9064** |
| set 2-nonAMP | AVP-HNCL | 1.0000 | 1.0000 | 1.0000 | 1.0000 |
| | AVP-IFT | 0.9934 | 0.9944 | 0.9925 | 0.9869 |
| | **our method** | **1.0000** | **1.0000** | **1.0000** | **1.0000** |

**Table5.** Performance comparison with other state-of-the-art methods on their respective benchmark datasets.

| Method | Acc | Sen | Spe | Mcc | G-Mean |
|---|---|---|---|---|---|
| FFMAVP | 0.9437 | 0.8802 | 0.9610 | 0.8345 | / |
| **Our** | **0.9689** | **0.9522** | **0.9734** | **0.9099** | **0.9628** |
| AVPiden | 0.9116±0.0031 | 0.9388±0.0053 | 0.9044±0.0045 | / | 0.9214±0.0023 |
| **Our** | **0.9492±0.0019** | **0.9079±0.0026** | **0.9707±0.0052** | **0.8865±0.0041** | **0.9388±0.0036** |

To comprehensively assess the overall competitiveness of AVP-Pro in the second-stage functional subtype prediction task, we systematically compared it with AVP-HNCL and AVP-IFT, which share an identical subtype test set. First, in terms of the macro-average performance across all 14 classification tasks (6 viral families + 8 specific viruses) as shown in Table 6, AVP-Pro exhibited a substantial overall advantage. Our model achieved the highest average accuracy (Avg. ACC = 0.9656) and average Matthews Correlation Coefficient (Avg. MCC = 0.8490), outperforming both AVP-HNCL (0.9596 / 0.8078) and AVP-IFT (0.8837 / 0.5758). A high average specificity (Avg.SP) of 0.9880 further underscores its exceptional capability in excluding non-target peptides, demonstrating that the model maintains extremely high predictive precision even when handling sparse data.

In the specific viral family classification tasks as shown in Table 7, AVP-Pro comprehensively outperformed AVP-IFT across all six families. Compared with the more competitive AVP-HNCL, our model achieved an MCC of 0.8838 for Coronaviridae, which was significantly higher than the competitor's 0.8002, demonstrating the model's performance advantage in this critical task. Furthermore, the perfect specificity (SP = 1.0000) achieved for the Orthomyxoviridae and Paramyxoviridae families further validates the model's potential for extremely low false-positive rates in clinical screening scenarios. For virus-specific recognition tasks as shown in Table 8, AVP-Pro also exhibited equally pronounced advantages. AVP-Pro's MCC scores for key viruses including HCV, RSV, and SARS-CoV (0.8838, 0.8757, and 0.9023, respectively) significantly outperformed those of both baseline models. Although some sensitivity was sacrificed to attain high specificity (0.9741) in individual tasks such as HIV prediction, the overall lead in MCC strongly validates the model's superiority in comprehensive discriminative capability. Notably, the perfect prediction across all metrics (1.0000) for HPIV3 provides compelling evidence for the model's high precision in identifying specific viral targets.

Additionally, to enable a fully fair comparison between AVP-Pro and state-of-the-art (SOTA) multi-classification methods for the second-stage functional subclass prediction task, we conducted fine-tuning experiments using the official best-fold partitioning scheme provided by FFMAVP. We compared the experimental results of AVP-Pro on this test set with the official performance metrics reported by FFMAVP and AVPIden as shown in Tables 9 and 10. The results show that for the 6-class virus family classification task (Table 9), AVP-Fusion outperformed both FFMAVP and AVPIden across the key metrics of MacroP (0.8840) and MacroF (0.8629). This outcome demonstrates more balanced and comprehensively superior classification capabilities. Although AVPIden achieved a marginally higher MacroR value, its MacroP metric was excessively low. This observation indicates a "broad coverage" prediction tendency that comes at the cost of accuracy, a trade-off that may lead to elevated false positive rates in practical screening scenarios.This trend became even more pronounced in the more challenging 8-class specific virus classification task (Table 10). AVP-Fusion once again delivered optimal performance with a MacroP of 0.8798 and a MacroF of 0.8774. Notably, as the number of classification categories increased, this method also exhibited strong competitiveness in the MacroR metric (0.8772). It significantly narrowed the gap with AVPIden (0.8962) while maintaining a substantial lead in MacroP. This finding confirms that AVP-Fusion retains exceptional precision and robustness when addressing finer-grained and more complex classification problems.

Table6. Overall Average Performance Comparison.

| Model | Avg ACC | Avg.MCC | Avg.SN | Avg.SP |
|---|---|---|---|---|
| AVP-HNCL | 0.9596 | 0.8078 | 0.9206 | 0.9676 |
| AVP-IFT | 0.8837 | 0.5758 | 0.9304 | 0.8837 |
| **our method** | **0.9656** | **0.8490** | 0.8083 | **0.9880** |

Table7. Summary of Performance on Independent Viral Family Datasets.

| Viral family | method | ACC | MCC | SN | SP |
|---|---|---|---|---|---|
| Coronaviridae | AVP-HNCL | 0.9700 | 0.8002 | 0.9189 | 0.9738 |
| | AVP-IFT | 0.8979 | 0.5709 | **0.9348** | 0.8952 |
| | **our method** | **0.9850** | **0.8838** | 0.8919 | **0.9919** |
| Flaviviridae | AVP-HNCL | **0.9456** | **0.8180** | 0.8469 | 0.9678 |
| | AVP-IFT | 0.8378 | 0.6406 | **0.9754** | 0.8070 |

|  | our method | 0.9417 | 0.7992 | 0.7071 | **0.9954** |
| --- | --- | --- | --- | --- | --- |
| Herpesviridae | AVP-HNCL | 0.9437 | **0.7582** | **0.9444** | 0.9436 |
|  | AVP-IFT | 0.8483 | 0.5199 | 0.8806 | 0.8447 |
|  | our method | **0.9455** | 0.6867 | 0.5690 | **0.9916** |
| Orthomyxoviridae | AVP-HNCL | 0.9587 | 0.6993 | **1.0000** | 0.9569 |
|  | AVP-IFT | 0.8036 | 0.3654 | 0.9655 | 0.7962 |
|  | our method | 0.9887 | 0.8318 | 0.7000 | **1.0000** |
| Paramyxoviridae | AVP-HNCL | **0.9925** | **0.9595** | 0.9636 | 0.9958 |
|  | AVP-IFT | 0.9625 | 0.8152 | 0.9118 | 0.9682 |
|  | our method | 0.9887 | 0.9418 | 0.8983 | **1.0000** |
| Retroviridae | AVP-HNCL | **0.9062** | 0.7985 | 0.8593 | **0.9341** |
|  | AVP-IFT | 0.9009 | **0.7954** | **0.9197** | 0.8897 |
|  | our method | 0.8872 | 0.7650 | 0.8194 | 0.9335 |

**Table8.** Summary of Performance on Independent Targeted Virus Datasets.

| Targeted virus | method | ACC | MCC | SN | SP |
| --- | --- | --- | --- | --- | --- |
| HCV | AVP-HNCL | **0.9719** | **0.9007** | **0.9432** | 0.9775 |
|  | AVP-IFT | 0.8442 | 0.5913 | 0.8818 | 0.8369 |
|  | our method | 0.9662 | 0.8838 | 0.8947 | **0.9817** |
| HIV | AVP-HNCL | **0.9193** | **0.8163** | **0.8736** | 0.9415 |
|  | AVP-IFT | 0.8818 | 0.7294 | 0.8241 | 0.9089 |
|  | our method | 0.9041 | 0.7876 | 0.7730 | **0.9741** |
| HPIV3 | AVP-HNCL | 0.9906 | 0.8697 | 0.9444 | 0.9922 |
|  | AVP-IFT | 0.9641 | 0.6786 | 0.9999 | 0.9629 |
|  | our method | **1.0000** | **1.0000** | **1.0000** | **1.0000** |
| HSV1 | AVP-HNCL | 0.9325 | 0.6656 | 0.8837 | 0.9367 |
|  | AVP-IFT | 0.8452 | 0.5161 | **0.9630** | 0.8350 |
|  | our method | **0.9624** | **0.7368** | 0.7209 | **0.9836** |
| INFVA | AVP-HNCL | 0.97 | 0.7412 | 0.9565 | 0.9706 |
|  | AVP-IFT | 0.8504 | 0.4131 | **0.9643** | 0.8454 |
|  | our method | **0.9906** | **0.8631** | 0.8000 | **0.9980** |
| RSV | AVP-HNCL | **0.9962** | **0.9556** | 0.9167 | **1.0000** |
|  | AVP-IFT | 0.9414 | 0.6388 | **0.9999** | 0.9386 |
|  | our method | 0.9906 | 0.8757 | 0.8182 | 0.9980 |
| SARS-CoV | AVP-HNCL | 0.9869 | 0.8704 | **0.8929** | 0.9921 |
|  | AVP-IFT | 0.8981 | 0.5005 | 0.8857 | 0.8987 |
|  | our method | **0.9906** | **0.9023** | 0.8571 | **0.9980** |
| FIV | AVP-HNCL | 0.9606 | 0.6560 | 0.9048 | 0.9629 |
|  | AVP-IFT | 0.8559 | 0.3863 | 0.9200 | 0.8534 |
|  | our method | **0.9868** | **0.7784** | 0.6667 | **0.9981** |

**Table9.** Performance comparison on the Stage2 virus-family classification benchmark (6 class).

| Method | MacroP | MacroR | MacroF |
| --- | --- | --- | --- |
| FFMAVP | 0.8712 | 0.8311 | 0.8481 |
| AVPIden | 0.6925 | **0.8675** | 0.7508 |
| **our method** | **0.8840** | 0.8452 | **0.8629** |

**Table10.** Performance comparison on the Stage2 target-virus classification benchmark (8-class).

| Method | MacroP | MacroR | MacroF |
| --- | --- | --- | --- |
| FFMAVP | 0.8724 | 0.8436 | 0.8551 |
| AVPIden | 0.6906 | **0.8962** | 0.7517 |
| **our method** | **0.8798** | 0.8772 | **0.8774** |

# Overview of amino acid distributions

To explore the universal amino acid composition characteristics of antiviral peptides (AVPs), we constructed a large-scale comprehensive dataset. This dataset integrates the Set 1-nonAVP dataset utilized for first-stage training in the present study, as well as the original first-stage datasets from the FFMAVP and AVPIden models, thereby maximizing the coverage of AVP sequence diversity.We compared the amino acid composition differences between AVPs and non-AVPs in the comprehensive dataset, with results as shown in Figure 1A. The analysis revealed a distinct preference for hydrophobic amino acids in AVP samples. Specifically, leucine (L) is extremely abundant in AVPs, and its abundance—along with that of isoleucine (I), valine (V), and tryptophan (W)—is significantly higher than that in non-AVP samples ($p < 0.001$). This high hydrophobicity may facilitate the insertion of AVPs into viral envelopes or their interaction with viral proteins. In terms of charge properties, although lysine (K) is relatively prevalent in both sample types, its proportion in AVPs remains significantly higher than that in non-AVP samples ($p < 0.001$). Combined with the similarly significant enrichment of arginine (R, $p < 0.01$) in AVPs, this indicates that AVPs as a whole carry a stronger positive charge, which favors binding to viral surfaces via electrostatic attraction.Conversely, the glycine (G) content in non-AVP samples is significantly higher than that in AVPs ($p < 0.001$). The reduction of glycine in AVPs suggests that these peptides may require a more rigid structure to sustain their functionality. Furthermore, the significant enrichment of cysteine (C) in AVPs ($p < 0.001$) may indicate that disulfide bonds play a crucial role in maintaining the active conformation of AVPs. In summary, high hydrophobicity, strong positive charge, and specific structural stability may represent the key structural requirements for AVPs to exert their antiviral activity.

To thoroughly investigate the functional specificity of distinct AVP subclasses, we performed a detailed analysis of amino acid composition variations across subclasses based on a high-quality Phase II dataset covering 6 viral families and 8 specific viruses. The heatmap in Figure 1B provides a visual illustration of amino acid enrichment (red) or depletion (blue) in each subclass relative to the universal AVP background. The results reveal substantial heterogeneity among AVPs targeting different viruses.First, AVPs targeting Paramyxoviridae and its associated viruses HPIV3 and RSV exhibit marked serine (S) enrichment. This likely correlates with serine-mediated hydrogen bond formation, which facilitates specific binding between the peptide and viral surface receptors. Notably, anti-HPIV3 peptides also display an extreme depletion of cysteine (C), suggesting that their structural stability may rely more on non-covalent interactions rather than disulfide bonds. Second, AVPs targeting Flaviviridae and the hepatitis C virus (HCV) show significant increases in valine (V) and lysine (K) abundances. This synergistic enrichment pattern of hydrophobicity and positive charge may reflect the requirement for specific electrostatic and hydrophobic mechanisms to target the unique envelope structures of Flaviviridae. In contrast, AVPs targeting Herpesviridae specifically exhibit arginine (R) enrichment, further underscoring the critical role of positive charge in distinct viral recognition processes.Additionally, we observed distinct clustering phenomena. For instance, the Retroviridae family and its member virus HIV exhibit highly similar patterns on the heatmap, without displaying extreme enrichment of any single amino acid. This suggests that anti-retroviral peptides may employ more diverse or conserved mechanisms of action. In contrast, anti-FIV peptides specifically enrich tryptophan (W), an aromatic amino acid typically crucial for maintaining peptide spatial structure and membrane insertion capability.

In summary, distinct AVP subclasses exhibit unique fingerprint-like characteristics in their amino acid composition. These characteristics not only complement the details overlooked in the first-phase general analysis—such as the critical role of arginine in specific subclasses—but also strongly validate the necessity of our refined subclass prediction approach implemented in the second phase.

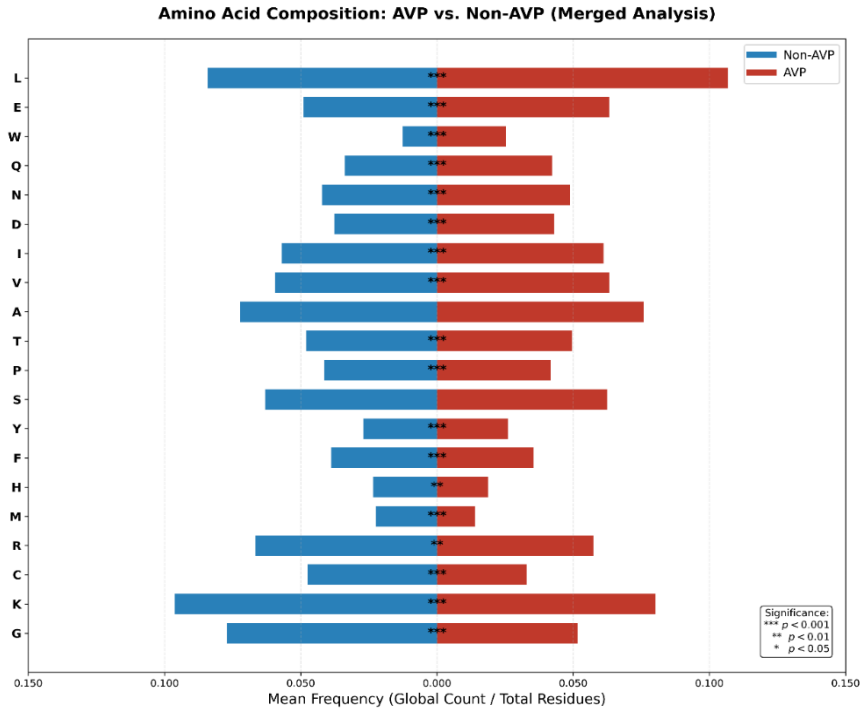

**Figure 1A**. Amino Acid Composition Analysis of AVPs and Non-AVPs in the Stage 1 Dataset. The butterfly chart compares the mean frequencies of the 20 standard amino acids in antiviral peptides (AVPs, red bars) versus non-antiviral peptides (Non-AVPs, blue bars). Bar length represents the average frequency within each category. Asterisks along the center line indicate the statistical significance of the differences, determined by a two-sample t-test (***: $p < 0.001$; **: $p < 0.01$; *: $p < 0.05$).

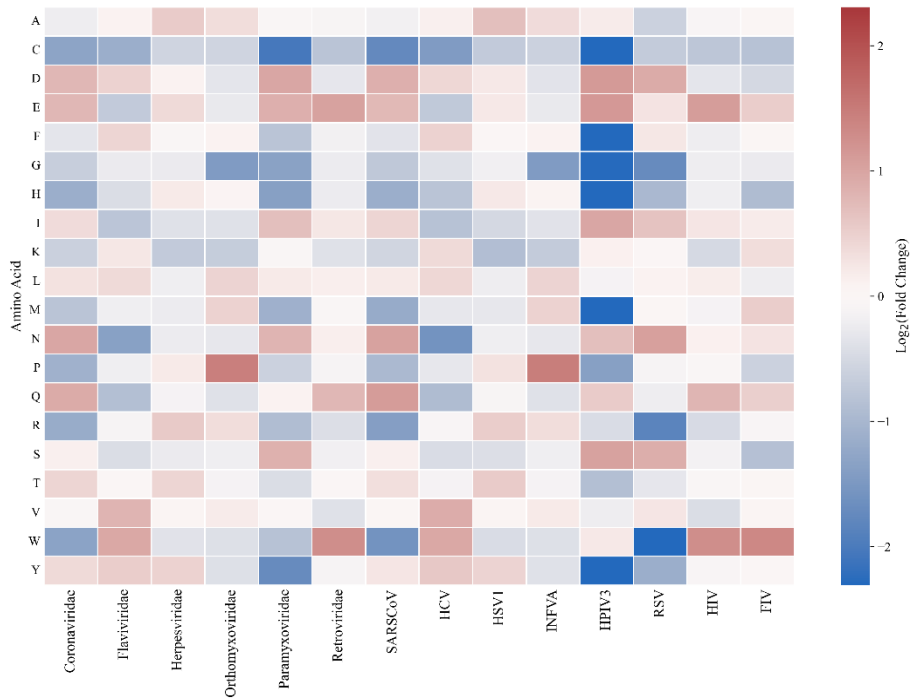

**Figure 1B**. Heatmap of Amino Acid Composition Specificity across AVP Subclasses. The heatmap displays the $\mathrm{Log}_2$ fold change of the mean frequency of each amino acid in the 14 functional subclasses relative to the general AVP background from Stage 1. Red cells indicate specific enrichment of an amino acid in a subclass, while blue cells indicate specific depletion. Subclasses (columns) are arranged logically by viral family followed by specific viruses, and amino acids (rows) are listed alphabetically for clarity.

# Ablation Study
## Validating Architectural Effectiveness

To systematically evaluate the contribution of each component within the AVP-Pro framework, we conducted a comprehensive ablation study. We compared the full model against two key variants. (1) Baseline, a foundational model that only incorporates ESM-2 and 10 handcrafted features into a multi-layer perceptron (MLP), without CNNs, BiLSTMs, or attention mechanisms. (2) No_Attention, a variant with only the hierarchical attention mechanism removed. The radar chart in Figure 2A provides a visual illustration of the multi-dimensional performance differences across models on five core metrics.Although the Baseline model possesses a certain predictive foundation owing to the powerful representation capability of the pre-trained language model, its performance on discriminative metrics (MCC = 0.858) is limited, indicating that simple feature concatenation struggles to capture deep sequential patterns. The No_Attention variant, which incorporates a dual-channel architecture, improves sensitivity (SN) but sacrifices specificity (SP) to a certain extent, indicating insufficient refinement of the decision boundary. In contrast, the full AVP-Pro model achieves a remarkable performance leap. Its MCC reaches 0.906 and ACC reaches 0.953, both substantially outperforming the Baseline and No_Attention variants. The maximum envelope area formed by AVP-Pro in the radar chart intuitively demonstrates its ability to maintain high sensitivity while substantially reducing the false positive rate (with SP as high as 0.942). This strongly validates that the integration of the hierarchical attention fusion architecture with the OHEM strategy significantly enhances the robustness of feature extraction.

To further uncover the underlying mechanism driving this performance improvement, we visualized the feature space of test set samples using Uniform Manifold Approximation and Projection (UMAP)[32] (Figure 2B). In the Baseline (left panel), AVP and non-AVP samples exhibit distinct overlapping regions with blurred decision boundaries. In contrast, the AVP-Pro model (right panel) exhibits a highly structured manifold distribution in the feature space. The two sample classes are clearly pushed to opposite ends, forming well-demarcated clusters. This significant improvement in feature representation geometrically confirms that our model successfully learned more discriminative sequence features, thereby underpinning the improvements in the aforementioned quantitative metrics.

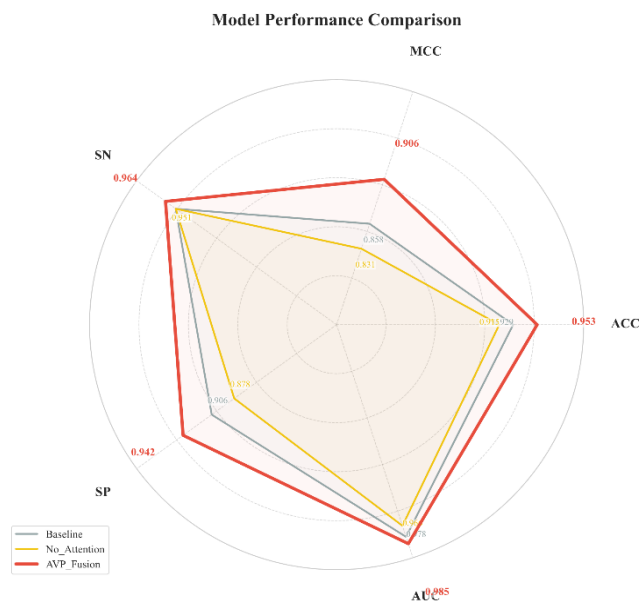

**Figure 2A**. Multi-Metric Performance Comparison (Radar Chart).

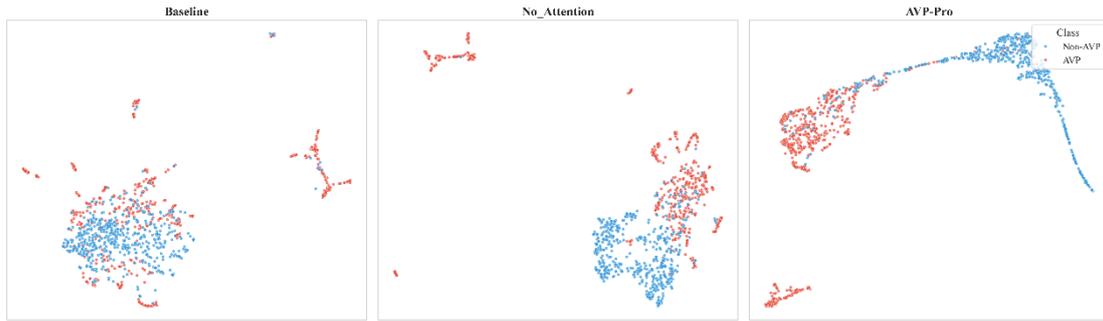

**Figure 2B**. Visualization of Feature Space Evolution via UMAP.

Furthermore, to elucidate how the Online Hard Negative Mining (OHEM) strategy optimizes the feature space during training, we monitored the dynamic evolution of the loss function and pairwise similarity. As shown in Figure 3, the training process exhibits a distinct two-stage optimization pattern. The gray dashed line indicates the steady decline of contrastive loss, confirming the model's convergence. More importantly, the solid lines reveal the operational mechanism of OHEM. The similarity of positive pairs (red) increases rapidly and stabilizes around 0.9, indicating that the model successfully aligns the representations of original AVPs and their augmented samples. Crucially, the similarity of hard negative pairs (blue) remains relatively high (> 0.6) in the initial phase, reflecting the model's initial difficulty in distinguishing AVPs from "hard" non-AVPs (e.g., peptides with similar amino acid compositions). However, as training progresses, this similarity decays sharply to below 0.2. This trend confirms that the OHEM mechanism actively "pushes away" the most deceptive negative samples from the positive sample cluster, thereby enforcing a larger margin at the decision boundary. This dynamic "pull-push" process is precisely the fundamental reason for the high specificity observed in our ablation experiments.

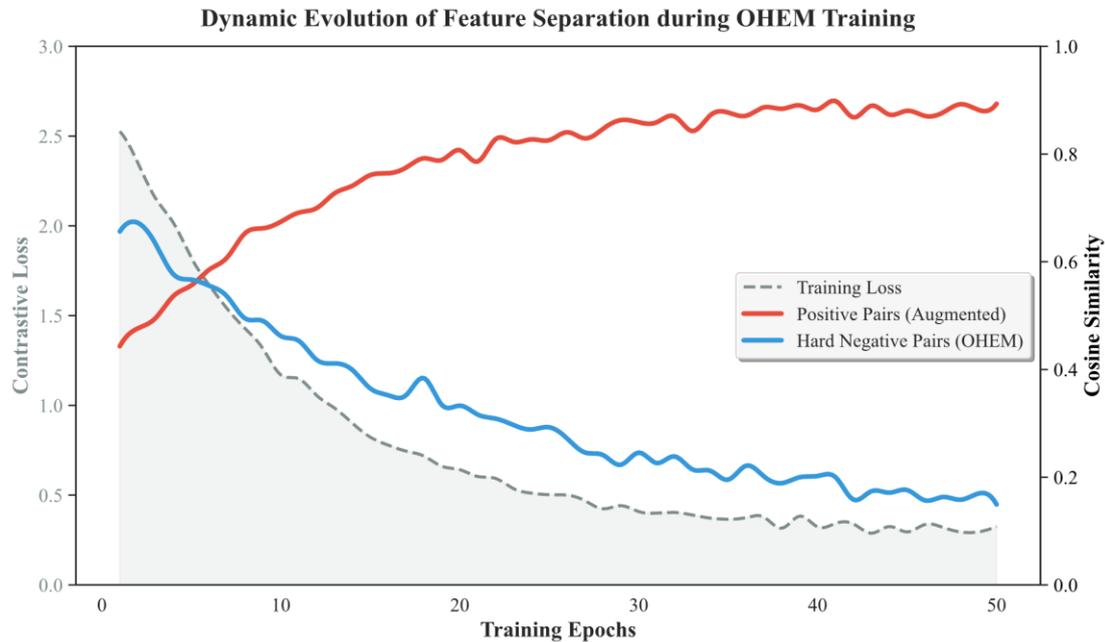

**Figure 3**. Dynamic Evolution of Feature Separation during OHEM Training. The horizontal axis represents training epochs. The left vertical axis displays the Contrastive Loss value, reflecting the optimization progress, and the right vertical axis represents the Cosine Similarity between feature vectors, ranging from 0 to 1. This plot tracks training dynamics over 50 epochs.

# Interpretability of the Adaptive Gating Mechanism

One of the core innovations of AVP-Pro resides in its adaptive gating network, a mechanism that dynamically adjusts feature weights based on the characteristics of input sequences. To investigate the operational mode of this mechanism, we statistically analyzed the distribution of gating weights ($\lambda$) across the test set (Figure 4). The results reveal that λ exhibits a bimodal distribution pattern with significant biological implications.For AVP samples (blue), the weights predominantly cluster around zero, indicating that the model relies heavily on global contextual features extracted by BiLSTM when recognizing antiviral peptides. This implies that the realization of antiviral functionality may not depend solely on specific local amino acid segments but rather on the overall sequence patterns and potential spatial structural features of the peptide. In contrast, for non-AVP samples (orange), the weights shift toward 1, indicating that the model prefers to utilize local motif features extracted by CNN. This may suggest that the model rapidly excludes negative samples by recognizing specific "inactive local fragments." This differentiated feature processing strategy explains how AVP-Pro achieves both sensitivity and specificity in complex samples.

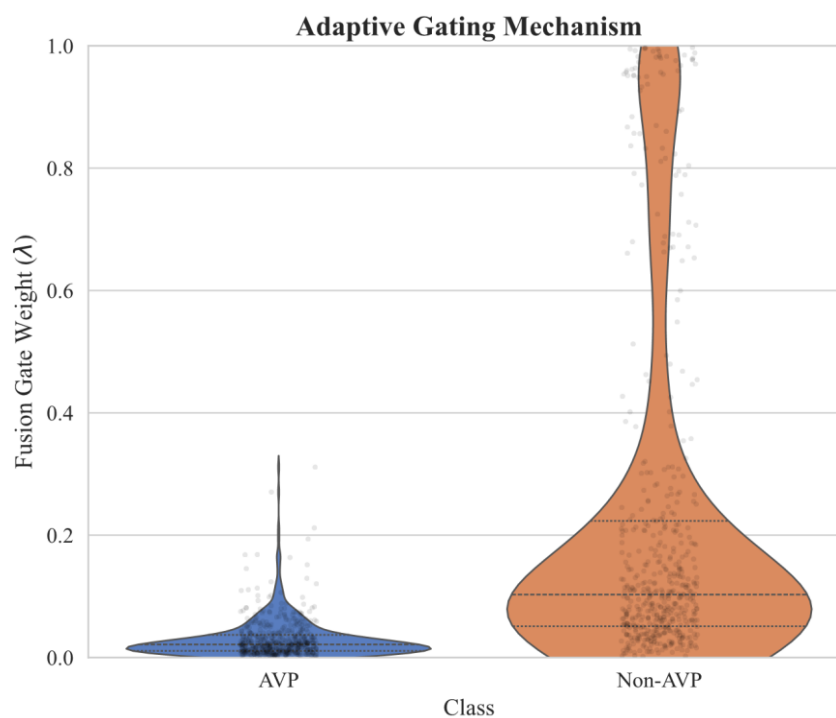

**Figure4**.Distribution Analysis of Adaptive Fusion Gating Weights.

## Capturing and Validating Biological Features

To verify whether the model has learned genuine biological patterns rather than data noise, we first analyzed the impact of data augmentation strategies. The t-SNE visualization [33] in Figure 5A shows that augmented samples generated under the guidance of the BLOSUM62 matrix (green) cluster closely around the original samples (red) in the feature space, maintaining strong semantic consistency. In contrast, randomly mutated samples (gray) exhibit divergence and semantic drift in the feature space. This confirms that data augmentation strategies incorporating evolutionary biological information can effectively enrich the training manifold while preserving the key functional features of peptide segments.

To visually interpret the model's decision-making rationale, we took the classic antiviral peptide Indolicidin [34] as a case study, extracted the raw weights from the self-attention layers of both the CNN and BiLSTM branches during model inference, and plotted the attention distribution maps after Min-Max normalization (Figure 5B). The analysis reveals that the blue solid line (CNN attention), which represents local features, exhibits significant peaks in the tryptophan (W) and proline (P) enrichment regions in the middle of the sequence, as well as at the C-terminal arginine (R) position. This distribution pattern is highly consistent with the well-established antiviral mechanism of Indolicidin, wherein the positively charged C-terminal region mediates the initial "adsorption" onto the viral surface via electrostatic interactions, while the central hydrophobic region rich in tryptophan drives subsequent membrane "insertion" and structural disruption. In contrast, the high-weight regions of the red dashed line (BiLSTM attention), which represents global features, are primarily concentrated at the N-terminus, indicating that this branch focuses on capturing the initial sequence signals and overall contextual dependencies. This complementary attention pattern strongly validates that AVP-Pro effectively integrates the microscopic recognition of key functional sites (via CNN) with the macroscopic regulation of overall sequence patterns (via BiLSTM), endowing the model with exceptional biological interpretability.

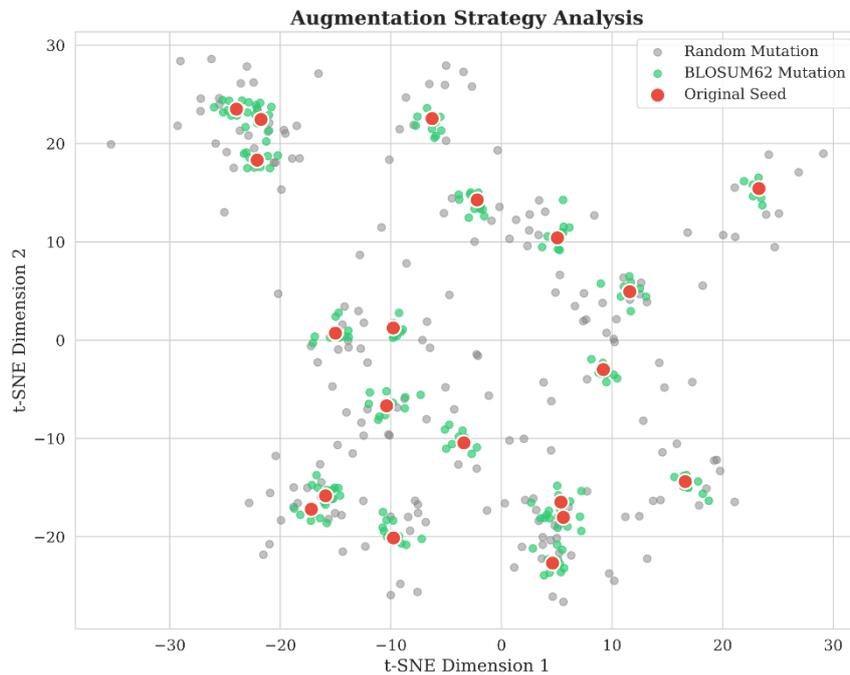

**Figure 5A**.The horizontal and vertical axes represent the two dimensions after t-SNE reduction, reflecting the relative distances of samples in the feature space.

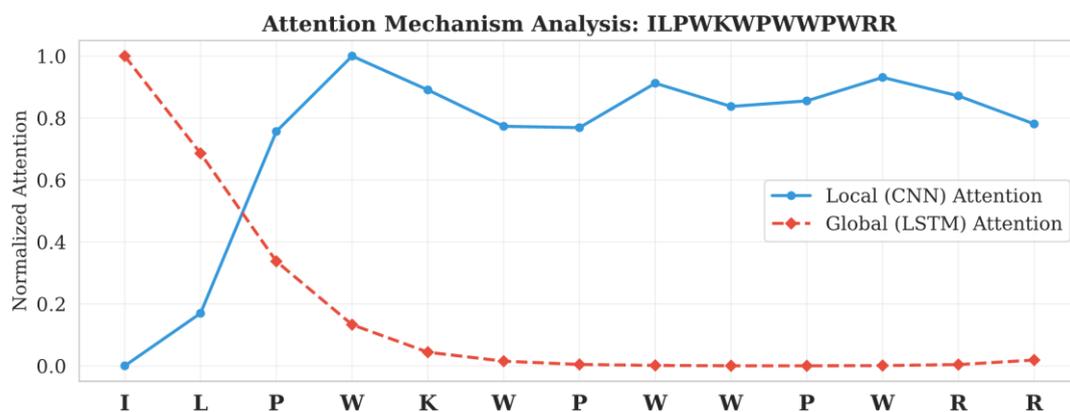

**Figure 5B**. Attention Weight Distribution Plot of Indolicidin. The horizontal axis represents the amino acid positions of the peptide sequence (ILPWKWPWWPWRR), and the vertical axis represents the normalized attention weight values (ranging from 0 to 1).

# Efficacy of Transfer Learning Strategy on Small-Sample Datasets

Given the paucity of annotated data for specific viral families (e.g., Coronaviridae), direct training of deep learning models from scratch often results in inadequate feature extraction and constrained generalization capability. To validate the necessity of our two-stage transfer learning strategy, we conducted comparative experiments on the Coronaviridae test set. Figure 6 illustrates the substantial performance divergence between the two approaches. As shown in the left bar chart, the model trained from scratch exhibits suboptimal performance on the discriminative metric MCC ($\approx 0.58$), indicating its failure to effectively distinguish specific antiviral targets from background noise. In stark contrast, transfer learning—where model weights are initialized using parameters pre-trained on the large-scale Stage 1 dataset—boosts the MCC score to above 0.88 and achieves an accuracy of 0.985, demonstrating the critical supportive role of pre-trained knowledge for downstream tasks. UMAP projections further elucidate the underlying mechanism. The feature space generated by the scratch-trained model (middle panel) exhibits a high degree of disorganization, with AVP (red) and non-AVP (blue) samples heavily intermingled, reflecting the model's inability to establish clear decision boundaries. In contrast, the transfer learning model (right panel) constructs a highly structured feature manifold. Within this space, anti-coronavirus peptides (Anti-CoV) no longer appear as scattered points but instead cluster into a compact, distinct island-like cluster, which is clearly segregated from the non-target background. This confirms that the universal antiviral features captured in Stage 1 were successfully adapted for the identification of specific coronavirus inhibitors, significantly mitigating the bottleneck caused by data scarcity.

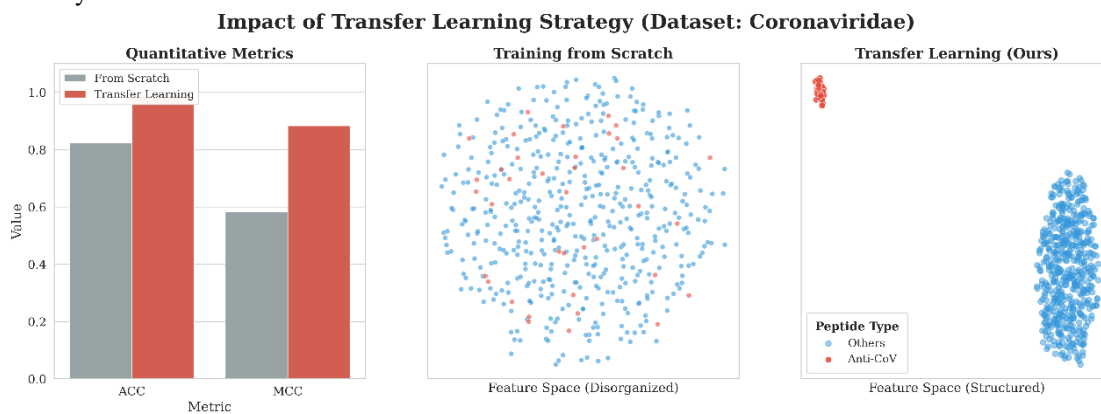

**Figure 6**. Validation of Transfer Learning Strategy on the Coronaviridae Dataset. The left panel shows the quantitative performance comparison. The middle panel displays the UMAP visualization of the model trained from scratch. The right panel presents the UMAP visualization of the transfer learning model.

# MATERIALS AND METHODS
## Dataset preparation

This study aims to develop an advanced computational framework for the prediction of antiviral peptides (AVPs) and their functional subclasses. To ensure the fairness of model performance evaluation and enable direct comparison with SOTA methods, we adopted a benchmark dataset constructed and publicly released by the recently proposed AVP-HNCL model developed by Li et al. This dataset architecture consists of two stages, dedicated to universal AVP recognition and functional subclass prediction, respectively.

For the universal AVP recognition task in the first stage, this study utilized two large-scale datasets, namely Set1-nonAVP and Set2-nonAMP. Both datasets share the same set of 2,662 experimentally validated AVP sequences as positive samples, which were extracted and consolidated from multiple authoritative bioactive peptide databases, including AVPdb[35], dbAMP[36], DRAMP[37], DBAASP[38], and HIPdb[39]. However, the key distinction between the two datasets lies in the composition of negative samples. The negative samples of Set1-nonAVP were derived from entries excluded for antiviral activity annotations in dbAMP, DBAASP, and DRAMP, while those of Set2-nonAMP were obtained from the UniProt database[40] and filtered by excluding a series of bioactivity-related keywords.To eliminate sequence redundancy, all sequences were clustered and deduplicated using the Cluster Database at High Identity with Tolerance (CD-HIT) tool, with a similarity threshold set at 40%[41]. This processing resulted in a class-balanced dataset, which was then strictly split into a training set and a test set at a 4:1 ratio. The detailed sample distributions of all datasets used in the first-stage task are summarized in Table 11.

For the functional subclass prediction task in the second stage, we similarly adopted the classification criteria from the HNCL dataset. AVP sequences were classified into six major viral families (Coronaviridae, Retroviridae, Flaviviridae, Orthomyxoviridae, Paramyxoviridae, and Herpesviridae) and eight specific viruses (e.g., Human Immunodeficiency Virus (HIV), Hepatitis C Virus (HCV)) based on their target specificity, where the selection of subclasses adheres to strict data volume criteria: each viral family must contain at least 100 sequence records, and each specific virus must have no fewer than 80 records.To rigorously avoid data leakage during the transfer learning process in the second stage, these 14 subclass datasets were constructed independently of the first-stage datasets and similarly split into training and testing sets at a 4:1 ratio for model fine-tuning and evaluation. Detailed information on these subclass datasets is presented in Tables 12 and 13. The adoption of this comprehensive set of publicly available and standardized datasets establishes a fair and robust foundation for comparing the performance of our AVP-Pro model with SOTA methods.

Table 11. Summary of the First-Stage Data Set.

| Dataset | training/test sets | positive samples | negative samples |
|---|---|---|---|
| Set 1-nonAVP | training | 2129 | 2129 |
| | test | 553 | 553 |
| Set 2-nonAMP | training | 2129 | 2129 |
| | test | 553 | 553 |

Table 12. Overview of multifunctional classified datasets-viral Families.

| viral family | Coronaviridae | Retroviridae | Herpesviridae | Paramyxoviridae | Orthomyxoviridae | Flaviviridae |
|---|---|---|---|---|---|---|
| Positive samples | 184 | 995 | 267 | 272 | 113 | 489 |
| Negative samples | 2478 | 1667 | 2395 | 2390 | 2549 | 2173 |

Table 13. Overview of multifunctional classified datasets-targeted Viruses.

| Targeted virus | FIV | HIV | HCV | HPIV3 | HSV1 | INFVA | RSV | SARS-CoV |
|---|---|---|---|---|---|---|---|---|
| Positive samples | 101 | 867 | 438 | 87 | 213 | 112 | 119 | 137 |
| Negative samples | 2561 | 1795 | 2224 | 2575 | 2449 | 2550 | 2543 | 2525 |

# Framework of the Proposed Model
## Sequence encodings

To comprehensively extract information from peptide sequences, we constructed a multimodal feature representation scheme. This scheme aims to simultaneously capture both the deep contextual semantic information and classical biophysical-chemical properties of peptide sequences, providing an information-rich and highly complementary input for subsequent deep learning models. Our feature engineering primarily consists of two components. These include deep representations based on large-scale pre-trained language models, and traditional feature descriptors derived from multiple sequence encoding algorithms.

First, we employed the protein pre-trained language model ESM2 (esm2_t30_150M_UR50D) to generate deep contextual embeddings for sequences[42]. ESM2 is a masked language model based on the Transformer architecture. Through self-supervised learning on billions of protein sequences, it gains profound insights into the "linguistic" patterns of proteins. We input each peptide sequence into the pre-trained ESM2 model and extracted the hidden state from its final layer as the sequence representation. This approach generates a context-sensitive, high-dimensional vector for each amino acid in the sequence, effectively capturing long-range dependencies and implicit evolutionary information between amino acids. This transfer learning strategy transfers knowledge learned from vast general protein data to the specific AVP prediction task.

Second, to supplement the classical global and local biochemical information that may be underrepresented by the ESM2 model, we additionally computed a series of traditional yet efficient feature descriptors, specifically encompassing the following ten distinct encoding methods:

(1) Basic Composition Features. Amino Acid Composition (AAC) and Dipeptide Composition (DPC)[43]. These features provide the most fundamental structural information of peptide sequences, where AAC and DPC respectively quantify the frequency of individual amino acids and adjacent amino acid pairs.

(2) Sequence Order and Association Features. K-spaced Amino Acid Group Pair Composition (CKSAAGP)[44], DistancePair[45], Pseudo-Amino Acid Composition (PAAC)[46], and Quasi-Sequence Order (QSOrder)[47]. To capture information beyond simple adjacency, we computed multiple sequence-order-related features. CKSAAGP and DistancePair respectively count the frequency of amino acid pairs separated by fixed or variable-length intervals. PAAC and QSOrder incorporate positional information and long-range association effects into feature vectors by introducing sequence-order-dependent factors.

(3) Physicochemical Properties and Grouping Pattern Features. Z-Scale [48, 49] and Generalized Topological Properties (GTPC)[50]. These features emphasize the chemical essence of amino acids. Z-Scale maps each amino acid into a five-dimensional vector space describing key physicochemical properties such as hydrophobicity and volume. GTPC first groups amino acids based on their physicochemical properties, then calculates the frequency of tripeptide group compositions to capture local chemical environments and patterns within sequences.

(4) Fundamental Encoding and Statistical Deviation Features. Binary Encoding and Dipeptide Deviation (DDE)[51]. Binary Encoding provides a unique one-hot representation for each amino acid. Additionally, we computed DDE, a feature that reveals potentially evolutionarily or structurally significant sequence signals by comparing the actual occurrence frequency of dipeptides against their theoretical expected frequency.

Through these steps, each peptide sequence is transformed into two distinct feature modalities. First, a dynamic, context-dependent deep embedding matrix derived from ESM. Second, a static, global attribute feature vector concatenating all ten traditional descriptors. These two features will be effectively fused within our subsequent model architecture, jointly driving the predictions of the AVP-Pro model.

## Data Augmentation

To construct high-quality positive sample pairs for contrastive learning tasks, we designed and implemented an innovative data augmentation strategy. Unlike conventional methods that perform random perturbations at the feature level, we opted to operate directly on the raw amino acid sequences. More importantly, our augmentation approach is not entirely stochastic.Instead, it incorporates biological prior knowledge, aiming to generate augmented sequences that are semantically similar to the original sequences yet structurally diverse.

Our data augmentation workflow consists of three core stages, namely segmentation, multi-step augmentation, and recombination. First, each input peptide sequence is randomly split into N fragments of approximately equal length. Subsequently, these fragments undergo iterative augmentation over a cycle of M steps. Within each step, one of three distinct augmentation strategies is alternately applied to the fragments based on their indices, that is BLOSUM62-based mutation, random insertion, or random deletion.

Among these strategies, BLOSUM62-based mutation constitutes the core of our approach. Traditional random mutations tend to generate biologically implausible sequences, thereby introducing extraneous noise into the training pipeline. To circumvent this limitation, we leveraged the BLOSUM62 substitution matrix, which encapsulates probabilistic information regarding amino acid substitutions during the natural evolution of proteins. For each amino acid targeted for mutation within the sequence, our algorithm queries the BLOSUM62 matrix and selects the amino acid with the highest substitution score that is distinct from the original residue for replacement. This "second-best" mutation strategy ensures that the augmentation operation simulates point mutations while maximally preserving the biochemical properties and structural stability of the original sequence.

The two additional augmentation strategies—random insertion and random deletion—involve introducing new amino acids into or removing existing residues from the fragments with a predefined probability, which simulates genetic insertion and deletion events and further enhances data diversity. After all augmentation steps are completed, all modified fragments are reassembled into a full-length augmented sequence. By virtue of this augmentation strategy that integrates segmented processing, multi-step iteration, and biologically informed guidance, we are able to generate positive samples that are functionally highly correlated with the original sequences yet exhibit divergent sequence patterns. This provides high-quality input for subsequent contrastive learning and consistency regularization tasks, thereby significantly enhancing the generalization capability and robustness of the model.

## Feature Engineering

To efficiently identify antiviral peptides (AVPs) and accurately predict their functional subclasses, we propose a novel deep learning framework named AVP-Pro. The core design philosophy of this framework lies in two key steps. First, hierarchically extracting multimodal features through parallel deep learning channels. Second, intelligently fusing these heterogeneous features using a novel adaptive gating mechanism. The entire model is an end-to-end network, with its overall architecture illustrated in Figure 7, primarily consisting of multimodal input fusion, parallel feature extraction, gated fusion, and a final classification module. By systematically integrating these 10 heterogeneous descriptors, we establish a panoramic view of peptide properties, compensating for the limitations of any single feature modality.

At the forefront of the model, we first concatenate the sequence embeddings from the ESM-2 model, which captures deep contextual semantics, with global feature vectors containing ten traditional biochemical descriptors. By extending the latter along the sequence length dimension and concatenating it with the former, we transform each peptide sequence into a robust multidimensional feature tensor. This approach effectively captures multidimensional information including the peptide's context, physicochemistry, and sequence patterns, providing a solid foundation for subsequent feature extraction. Next, the fused feature tensor is fed into a parallel feature extractor. This extractor is the core of the model, drawing upon and evolving the design concept of dual-channel networks. It is composed of a Convolutional Neural Network (CNN) branch and a Bidirectional Long Short-Term Memory (BiLSTM) branch operating in parallel, aiming to simultaneously capture local patterns and global dependencies within the sequence. The CNN branch is responsible for extracting local conserved motif information from the sequence. We employ a one-dimensional convolutional neural network (1D-CNN) to process the fused sequence representation and efficiently extract local features via convolutional kernels. For an input sample x, the feature $c_j^k$ generated by the k-th convolutional kernel at the j-th position can be expressed as follows.

$$c_j^k = f\left(\sum_{i=0}^{H-1} \mathbf{W}_i^k \cdot \mathbf{x}_{j+i} + b^k\right) \tag{1}$$

Where $f$ represents the ReLU activation function ($f(x) = \max(0, x)$), $H$ is the size of the convolutional kernel, $W^k$ is the weight tensor of the $k$-th kernel, $b^k$ is the corresponding bias term, and $x_{j+i}$ denotes the $i$-th feature vector within the window of length $H$ starting from position $j$ in the input sequence. By sliding the convolutional kernel across the entire sequence, we obtain a complete feature map. Compared to more complex convolutional methods, 1D-CNN offers a simpler structure with fewer parameters, significantly enhancing computational efficiency when processing sequence information. The BiLSTM branch is utilized to capture long-range dependencies between amino acids. As a special type of Recurrent Neural Network (RNN), BiLSTM, through its forward and backward propagation mechanisms, can simultaneously capture past and future contextual information in the sequence, thereby better handling long-term dependency issues and enhancing the model's contextual understanding capabilities. The core formulas of the LSTM unit are as follows.

$$f_t = \sigma(W_f \cdot [h_{t-1}, x_t] + b_f) \tag{2}$$

$$i_t = \sigma(W_i \cdot [h_{t-1}, x_t] + b_i) \tag{3}$$

$$\tilde{C}_t = \tanh(W_C \cdot [h_{t-1}, x_t] + b_C) \tag{4}$$

$$C_t = f_t \cdot C_{t-1} + i_t \cdot \tilde{C}_t \tag{5}$$

$$O_t = \sigma(W_0 \cdot [h_{t-1}, x_t] + b_0) \tag{6}$$

$$h_t = O_t \cdot \tanh(C_t) \tag{7}$$

The above system of equations precisely defines the state transition process of the LSTM unit at each time step t. This process is regulated by three gating mechanisms which are the forget gate $f_t$, the input gate $i_t$, and the output gate $O_t$. The cell state $C_t$ serves as the core memory unit of the network. Controlled by $f_t$ and $i_t$, it determines how much historical information to retain and how much new information represented by $C_t$ to incorporate. The hidden state $h_t$ is the output of the network at that time step, summarizing the sequence information at the current moment. This series of dynamic calculations is parameterized by weight matrices (e.g., $W_f$, $W_i$) and bias vectors (e.g., $b_f$, $b_i$) and utilizes the tanh function for non-linear transformation. In the bidirectional model (BiLSTM), the input sequence is processed in parallel by two independent LSTM networks—forward and backward—producing the forward hidden state $h_t^{forward}$ and the backward hidden state $h_t^{backward}$, respectively. Finally, we concatenate these two state vectors to generate a final representation rich in complete bidirectional contextual information for each position in the sequence. At the ends of both branches, we introduce self-attention mechanisms to dynamically identify and aggregate the feature information in the sequence that contributes most to the prediction, forming the respective vector representations $V_{cnn}$ and $V_{bilstm}$.

The key innovation of the AVP-Pro model lies in its Gated Fusion Network. To intelligently fuse the heterogeneous features extracted by the CNN and BiLSTM, we designed an adaptive gating mechanism. This network takes the output vectors $V_{cnn}$ and $V_{bilstm}$ from the two branches as input and learns a dynamic gating weight $\lambda$ through a small neural network. This weight is then used to perform a weighted sum of the two feature vectors, resulting in the final fused embedding representation $E_{final}$.

$$E_{final} = \lambda \cdot f_{match}(v_{cnn}) + (1-\lambda) \cdot v_{bilstm} \tag{8}$$

This gating mechanism allows the model to adaptively decide whether to prioritize local sequence motifs or global contextual dependencies based on the intrinsic characteristics of the input sequence itself, achieving more efficient and flexible feature fusion than simple concatenation or addition. Finally, this highly condensed fused embedding vector $E_{final}$ is fed into a Multilayer Perceptron (MLP) classifier to output the final prediction result.

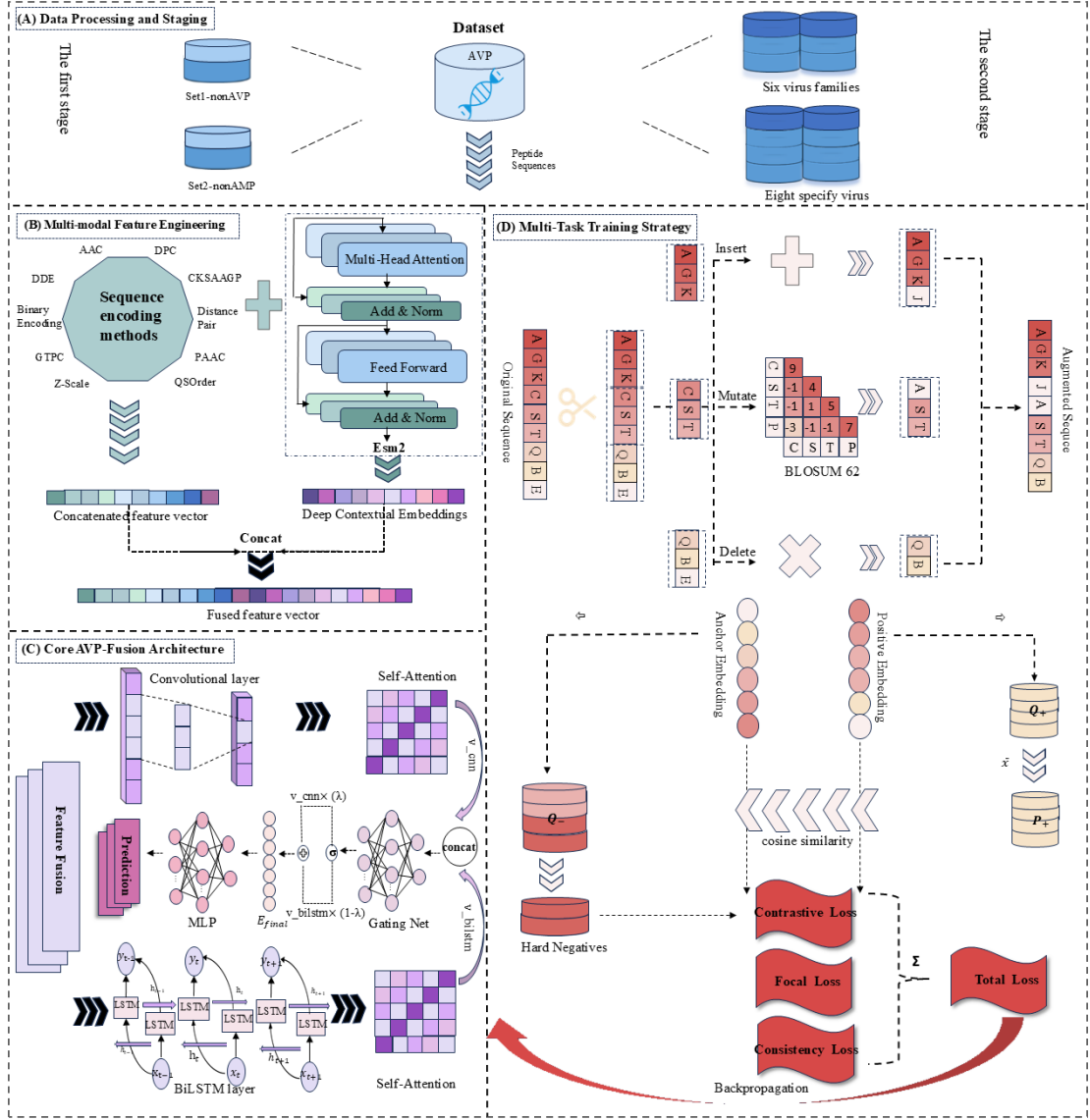

**Figure 7.** The overall framework of the proposed AVP-Pro model. This framework consists of four main modules, illustrating the complete workflow from data processing to model training. (A) Data Processing and Task Segmentation. This module outlines the data sources and the two-stage prediction pipeline. Stage 1 utilizes the Set1-nonAVP and Set2-nonAMP datasets for general AVP identification. Stage 2 employs transfer learning, building upon the knowledge from Stage 1, to predict functional subclasses across datasets for six viral families and eight specific viruses. (B) Multimodal Feature Engineering. Peptide sequences are processed through two parallel pathways to generate comprehensive feature representations. The upper pathway uses a pre-trained ESM-2 model to generate deep contextual embeddings, capturing semantic and evolutionary information. The lower pathway utilizes various sequence encoding methods to calculate ten traditional biochemical descriptors, providing a static, global perspective of the peptide. These two distinct sets of features are then concatenated to form a fused feature tensor. (C) Core AVP-Pro Architecture: This is the heart of our model. The fused feature tensor is fed into a parallel network comprising a CNN-Attention branch for capturing local motifs and a BiLSTM-Attention branch for modeling long-range dependencies. The outputs of these two branches, V_cnn and V_bilstm are intelligently integrated by a novel adaptive gating mechanism, which computes a dynamic weight λ to produce a final, highly condensed embedding vector (E_final). This embedding is then passed to a Multilayer Perceptron (MLP) classifier to generate the final prediction. (D) Multi-Task Training Strategy. The model is trained end-to-end using a composite loss function. This module details the key components of the training strategy: (i) a biologically informed augmentation scheme that generates positive sample pairs by applying mutations guided by the BLOSUM62 matrix, as well as probabilistic insertion and deletion operations on sequence segments. (ii) a novel prototype-based OHEM contrastive learning scheme, which mines hard negative samples from a dynamic negative queue (Q_-) based on a positive prototype (P_+) calculated from a positive sample queue (Q_+), and (iii) a multi-task objective consisting of Focal Loss for classification, Contrastive Loss for representation learning, and Consistency Loss for regularization. The total loss is then used to update model parameters via backpropagation.

## Contrastive learning

For the contrastive learning component, we did not rely on a single classification loss. Instead, by co-optimizing three complementary learning objectives, the model not only achieves accurate classification but also learns well-structured and robust feature representations. The final total loss function $L_{total}$ is thus defined as a weighted sum of three key loss terms.

$$L_{\text{total}} = \lambda_1 L_{\text{con}} + L_{\text{cls}} + \lambda_2 L_{\text{cons}} \tag{9}$$

Where $L_{con}$ is the contrastive loss, $L_{cls}$ is the primary classification loss, and $L_{cons}$ is the consistency regularization loss. $\lambda_1$ and $\lambda_2$ are hyperparameters used to balance the contributions of each task.

(1) Contrastive Loss Based on OHEM Queue. To guide the model in learning a semantically structured embedding space, we introduced a novel contrastive learning task. Its core objective is to pull similar sample pairs (positive pairs) closer together in the feature space while pushing dissimilar sample pairs (negative pairs) apart [52]. The construction of positive pairs is based on our proposed data augmentation strategy. For each positive sample $x_a$ (anchor), the sequence obtained after augmentation via BLOSUM62-based mutation, random insertion, and deletion is considered its positive sample $x_p$ (positive). The selection of negative sample pairs is a key innovation of our training strategy. Traditional contrastive learning typically employs in-batch random negative sampling, which is inefficient and may fail to provide sufficiently challenging negative samples. To overcome this limitation, we designed and implemented a queue-based Online Hard Negative Mining (OHEM) strategy. This strategy relies on a positive sample queue ($Q_+$) and a negative sample queue ($Q_-$). When selecting negative samples for a given anchor sample $x_a$, we do not simply calculate the individual similarity between the anchor and the negative samples. Instead, we adopt a difficulty measure with a more global perspective. Specifically, we first calculate the mean of all features in the positive sample queue $Q_+$ to form a dynamically updated positive prototype ($P_+$), which represents the feature center of the "standard AVP" as currently understood by the model. Next, we assess the "difficulty" of each sample in the negative sample queue $Q_-$ by calculating its cosine similarity with this positive prototype—the closer a negative sample is to the cluster center of positive samples in the feature space, the more confusing it is, and thus the higher its difficulty. The cosine similarity is calculated as follows.

$$\sin(x, y) = \frac{x \cdot y}{\| x \| \| y \|} \tag{10}$$

Where $x$ and $y$ represent the embedding vectors of the samples. Finally, based on these difficulty scores, we select a batch of the most challenging negative samples via weighted random sampling for the calculation of the contrastive loss. The mathematical expression for the contrastive loss function is:

$$L_{\text{con}} = -\frac{1}{N_a} \sum_{a=1}^{N_a} \log \frac{\exp\left(\frac{\sin(x_a, x_p)}{\tau}\right)}{\exp\left(\frac{\sin(x_a, x_p)}{\tau}\right) + \sum_{k=1}^{K} \exp\left(\frac{\sin(x_a, n_k)}{\tau}\right)} \tag{11}$$

Where, $N_a$ denotes the number of anchor samples in a batch, $x_a$ represents the anchor sample, $x_p$ denotes its corresponding positive sample, and $n_k$ represents the $k$-th hard negative sample. The function $\text{sim}(\cdot,\cdot)$ computes the cosine similarity between two vectors. $K$ denotes the number of hard negative samples associated with each anchor, while $\tau$ is a learnable temperature hyperparameter that adjusts the sharpness of the similarity distribution to regulate the learning difficulty. This loss function is designed to learn a more discriminative feature space by maximizing the similarity of positive sample pairs and minimizing the similarity with hard negative samples.

(2) Focal Classification Loss. Considering the prevalent class imbalance issue in the AVP dataset, we adopt Focal Loss [53] as the primary classification supervision signal. This choice is complementary to our OHEM-based contrastive learning strategy. While the OHEM strategy actively mines hard negative samples in the feature space during the representation learning phase, Focal Loss dynamically adjusts sample weights during the final classification decision phase, directing the model to focus more attention on hard-to-distinguish samples with low prediction probabilities. This dual focus on hard samples at both the representation and decision levels enables us to significantly improve the model's classification accuracy. Furthermore, the introduction of Focal Loss lays a methodological foundation for addressing the highly imbalanced functional subclass prediction task in the subsequent second stage. For a classification task with $C$ categories, its mathematical expression is given by:

$$L_{\text{focal}} = -\frac{1}{N}\sum_{i=1}^{N}\alpha_{y_i}(1-p_{i,y_i})^{\gamma}\log(p_{i,y_i}) \tag{12}$$

Where $N$ is the batch size, $y_i$ is the true label of sample $i$, and $p_{i,y_i}$ is the probability that the model predicts sample $i$ as the true class $y_i$. $\gamma$ is the focusing parameter used to adjust the weights of easy and hard samples, and $\alpha_{y_i}$ is the class weight parameter used to balance the importance of different classes. This loss function effectively improves the model's performance on imbalanced data.

(3) Consistency Regularization Loss. To further enhance the model's generalization capability and predictive robustness, we introduce Consistency Regularization Loss [54] as the third learning objective. This loss quantifies the consistency of predictions by calculating the Symmetric KL-Divergence between the predicted probability distributions $P$ and $P'$ of the original positive sample $x$ and its augmented version $x'$ at the classifier output as follows.

$$L_{\text{cons}} = \frac{1}{2}(D_{KL}(P(y|x)||P(y|x')) + D_{KL}(P(y|x')||P(y|x))) \tag{13}$$

Where $D_{KL}(\cdot||\cdot)$ is the standard Kullback-Leibler divergence. This loss term does not rely on true labels. By encouraging the model to learn a smooth decision surface, it effectively reduces the risk of overfitting. This design aims to address a specific challenge introduced by our advanced data augmentation strategy, whereby biological prior-based augmentation generates high-quality positive sample pairs, it may also induce the model to learn fragile surface features that are overly sensitive to minor sequence perturbations, rather than the intrinsic biological essence of AVP activity. To overcome this risk, we impose a strong consistency constraint at the model's decision level by minimizing the symmetric KL divergence between the prediction distributions of the original and augmented samples. This constraint, combined with contrastive learning that pulls sample distances closer in the representation layer, forms a dual safeguard, synergistically ensuring that the model not only learns structured feature representations but also makes robust and generalizable predictions based on them.

## Prediction Module

To obtain the final prediction result, the final embedding vector $E_{final}$, obtained from the gated fusion network, is fed into a Multilayer Perceptron (MLP) classifier. This classifier consists of multiple fully connected layers, batch normalization, and activation functions. Its core task is to map the high-dimensional feature representations learned by the model into the final class prediction space. The entire prediction process can be simplified as follows.

$$y_{\text{pred}} = \text{Classifier}(E_{\text{final}}) \tag{14}$$

Where $E_{final}$ is the final feature vector defined in Section 2.4.4, which integrates both local and global information. $y_{pred}$ represents the model's prediction for the input sample. For binary classification tasks, this output typically indicates the probability that the sample belongs to the positive class.

## Transfer Learning

Upon completion of the AVP universal recognition model construction, we proceeded to the core task of the second phase, which is predicting the functional subclasses of AVPs targeting specific viral families and species. Given that datasets for these functional subclasses typically exhibit small sample sizes and highly imbalanced class distributions, training deep learning models from scratch is highly susceptible to overfitting and suboptimal performance. To address this "few-shot learning" challenge, we adopted a transfer learning strategy, which aims to transfer the feature extraction capabilities acquired from the large-scale, universal dataset in Phase 1 to the specific subtasks in Phase 2.

Specifically, we first loaded the checkpoint of the AVP-Pro model that achieved the optimal performance during Phase 1 training. Subsequently, we initialized the parameters of its robust feature extraction network—including the multimodal input layer, parallel CNN and BiLSTM branches, and the gated fusion network—using the pre-trained weights. For each subclass prediction task, only the top-level classifier was replaced and randomly initialized. The entire model was then fine-tuned end-to-end on the corresponding subclass dataset. This approach enables the model to rapidly adapt to and learn the subtle features that distinguish specific functional subclasses using minimal labeled data, while retaining its capability of general peptide sequence representation.

To further enhance model stability and prediction accuracy during the fine-tuning phase, we employed a Test-Time Augmentation (TTA) strategy [55] at the final inference stage. This strategy applies multiple minor random mutations based on the BLOSUM62 matrix to each test sample, generating a series of semantically similar augmented samples. The final prediction probability is obtained by averaging the prediction results of the original sample and all augmented samples. This method effectively smooths the prediction outputs, reduces the random bias introduced by a single forward propagation, and thereby achieves more robust and reliable classification performance.

## Evaluation metrics

To comprehensively and objectively evaluate the performance of the AVP-Pro model in both first-stage and second-stage tasks, we adopted a series of standard evaluation metrics widely used in bioinformatics classification tasks. For binary classification tasks, we calculated Accuracy (ACC), Sensitivity (Sen), Specificity (Sep), Matthews Correlation Coefficient (MCC), F1-Score, and Geometric Mean (G-mean)[56]. Furthermore, to assess the model's overall performance across different thresholds, we also calculated the Area Under the Receiver Operating Characteristic Curve (AUROC) and the Area Under the Precision-Recall Curve (AUPRC). The calculation formulas for these metrics are as follows.

$$Accuracy = \frac{TP + TN}{TP + TN + FP + FN} \qquad (15)$$

$$Sensitivity = \frac{TP}{TP + FN} \qquad (16)$$

$$Specificity = \frac{TN}{TN + FP} \qquad (17)$$

$$MCC = \frac{TP \times TN - FP \times FN}{\sqrt{(TP + FP)(TP + FN)(TN + FP)(TN + FN)}} \qquad (18)$$

$$G\text{-}mean = \sqrt{Sensitivity \times Specificity} \qquad (19)$$

$$AUROC = \int_0^1 TPR(t) d(FPR(t)) \qquad (20)$$

where TPR (True Positive Rate) = TP / (TP + FN), FPR (False Positive Rate) = FP / (FP + TN), and t represents the classification threshold.

$$AUPRC = \int_0^1 Precision(r) dr \qquad (21)$$

To facilitate comparisons in the second phase, we also employed commonly used multi-class evaluation metrics, namely MacroP, MacroR, and MacroF. Their respective formulas are provided below.

$$MacroP = \frac{1}{c} \sum_{i=1}^{c} \frac{TP_i}{TP_i + FP_i} \qquad (22)$$

$$MacroR = \frac{1}{c} \sum_{i=1}^{c} \frac{TP_i}{TP_i + FN_i} \qquad (23)$$

$$MacroF = \frac{1}{c} \sum_{i=1}^{c} \frac{2 Precision_i * Recall_i}{Precision_i + Recall_i} \qquad (24)$$

where Precision = TP / (TP + FP), Recall = TP / (TP + FN), and r represents the recall rate. $Precision_i$ and $Recall_i$ represent the precision and the recall under class i . In the above formulas , TP,

TN, FP, and FN denote True Positives, True Negatives, False Positives, and False Negatives, respectively.

## Implementation details

Our proposed AVP-Pro model and all experiments were implemented based on the PyTorch (1.13.0 with CUDA 11.7) deep learning framework[57]. Model training and evaluation were conducted on Linux servers equipped with NVIDIA GPUs. To ensure the reproducibility of the study, we fixed the global random seed prior to the initiation of all experiments.

The input features of the model consist of two components. Deep contextual embeddings were generated using the pre-trained ESM-2 model loaded via the Transformers library (4.28.1). Traditional biochemical feature descriptors were primarily computed using iFeatureOmega (1.0.2)[58] and custom Python scripts. The core workflow for constructing and evaluating machine learning models relied on the Scikit-learn scientific computing library.

During the model training phase, we uniformly adopted the AdamW optimizer[59]. In the pre-training stage of Phase 1, the initial learning rate was set to 1.2e-4 with a weight decay of 1e-2, and a cosine annealing learning rate scheduler with a warm-up phase was employed. For the fine-tuning tasks in Phase 2, the learning rate was adjusted to 8.0e-5 and the weight decay was reduced to 0.0 to adapt to few-shot learning. The entire training process prevented overfitting by monitoring the performance on the validation set and adopting an early stopping strategy. Additionally, we utilized gradient accumulation techniques to stabilize mini-batch training. All critical hyperparameters—such as the dropout rate and contrastive loss weight—were determined in the Phase 1 task and maintained consistently across all fine-tuning tasks in Phase 2, ensuring the fairness and effectiveness of model transfer.

## Conclusion

Faced with the formidable challenges posed by viral mutations and drug resistance, the development of novel therapies based on AVPs has become an urgent need in the biopharmaceutical field. However, traditional experimental screening methods are not only costly but also time-consuming. In contrast, data-driven computational models can rapidly identify potential drug candidates from vast sequence libraries, significantly improving R&D efficiency. Against this backdrop, we developed the AVP-Pro framework, a two-stage deep learning architecture that not only achieves high-precision AVP identification but also enables accurate prediction of their functional subtypes targeting specific viruses. The outstanding performance of AVP-Pro is primarily attributed to four innovative design aspects.

First, we adopted a more biologically meaningful data augmentation strategy. By incorporating the BLOSUM62 substitution matrix, we simulated the natural evolutionary process of amino acids. Samples generated by this method not only enhance data diversity but also preserve the key biochemical properties of peptides, thereby significantly boosting the model's generalization capability.

Second, we designed an intelligent feature fusion architecture. This architecture integrates an adaptive gating mechanism that dynamically adjusts the weights of local features (extracted by CNN) and global features (extracted by BiLSTM) based on the specific sequence context. This mechanism enables the model to capture complex sequence patterns with greater flexibility.

Third, to address the challenge of distinguishing similar samples, we implemented an Online Hard Example Mining (OHEM)-based contrastive learning strategy. This strategy focuses on mining "hard samples" that are prone to misclassification for intensive training. Through this targeted learning, the model significantly sharpens its decision boundary, thereby effectively reducing the false positive rate and improving prediction specificity.

Fourth, AVP-Pro exhibits excellent interpretability. Visualization of the model's attention weights reveals that the model can precisely focus on key functional sites within peptides (e.g., hydrophobic cores or charged termini). This not only validates the reliability of the model's decisions but also provides intuitive insights for subsequent rational drug design.

Despite significant progress, the current model primarily relies on one-dimensional sequence features and still has room for improvement in predicting certain viral subfamilies with sparse data. Future work will focus on incorporating three-dimensional structural information (e.g., AlphaFold-predicted structures) and exploring active learning strategies to further enhance the model's robustness and applicability. In summary, AVP-Pro provides a powerful and reliable computational tool for the screening and development of antiviral drugs.

# Acknowledgements

We sincerely express our gratitude to all members of our research group for their valuable efforts and guidance, as well as to the reviewers for their constructive comments and suggestions.

# Funding

This work was supported by grants from the National Natural Science Foundation of China (No.32260154, 62562041).

# Data availability

Codes and datasets are available at https://github.com/wendy1031/AVP-Pro

# References

1.  Hollmann, A., et al., *Review of antiviral peptides for use against zoonotic and selected non-zoonotic viruses.* Peptides, 2021. **142**: p. 170570.
2.  Kumari, M., et al., *A critical overview of current progress for COVID-19: development of vaccines, antiviral drugs, and therapeutic antibodies.* J Biomed Sci, 2022. **29**(1): p. 68.
3.  Aylward, F.O. and M. Moniruzzaman, *Viral Complexity.* Biomolecules, 2022. **12**(8).
4.  Ngai, P.H., T.J.B. Ng, and C. Biology, *Phaseococcin, an antifungal protein with antiproliferative and anti-HIV-1 reverse transcriptase activities from small scarlet runner beans.* Biochemistry and Cell Biology 2005. **83**(2): p. 212-220.
5.  Quintero-Gil, C., et al., *In-silico design and molecular docking evaluation of peptides derivatives from bacteriocins and porcine beta defensin-2 as inhibitors of Hepatitis E virus capsid protein.* Virusdisease, 2017. **28**(3): p. 281-288.
6.  Qureshi, A., *A review on current status of antiviral peptides.* Discover Viruses, 2025. **2**(1).
7.  Su, S., et al., *A peptide-based HIV-1 fusion inhibitor with two tail-anchors and palmitic acid exhibits substantially improved in vitro and ex vivo anti-HIV-1 activity and prolonged in vivo half-life.* Molecules, 2019. **24**(6): p. 1134.
8.  Manzano-Robleda Mdel, C., et al., *Boceprevir and telaprevir for chronic genotype 1 hepatitis C virus infection. A systematic review and meta-analysis.* Ann Hepatol, 2015. **14**(1): p. 46-57.
9.  Yao, H., et al., *Molecular Architecture of the SARS-CoV-2 Virus.* Cell, 2020. **183**(3): p. 730-738 e13.
10. Lefin, N., et al., *Review and perspective on bioinformatics tools using machine learning and deep learning for predicting antiviral peptides.* Molecular diversity, 2024. **28**(4): p. 2365-2374.
11. Thakur, A., et al., *In pursuit of next-generation therapeutics: antimicrobial peptides against superbugs, their sources, mechanism of action, nanotechnology-based delivery, and clinical applications.* International journal of biological macromolecules, 2022. **218**: p. 135-156.
12. Chang, K.Y. and J.-R.J.P.o. Yang, *Analysis and prediction of highly effective antiviral peptides based on random forests.* PloS one, 2013. **8**(8): p. e70166.
13. Lissabet, J.F.B., et al., *AntiVPP 1.0: a portable tool for prediction of antiviral peptides.* Computers in biology and medicine, 2019. **107**: p. 127-130.
14. Schaduangrat, N., et al., *Meta-iAVP: a sequence-based meta-predictor for improving the prediction of antiviral peptides using effective feature representation.* International journal of molecular sciences, 2019. **20**(22): p. 5743.
15. Akbar, S., et al., *Prediction of antiviral peptides using transform evolutionary & SHAP analysis based descriptors by incorporation with ensemble learning strategy.* Chemom Intel Lab Syst, 2022. **230**: p. 104682.
16. Li, J., et al., *DeepAVP: a dual-channel deep neural network for identifying variable-length*

16. *antiviral peptides.* IEEE Journal of Biomedical and Health Informatics, 2020. **24**(10): p. 3012-3019.
17. Li, Z., et al., *A Survey of Convolutional Neural Networks: Analysis, Applications, and Prospects.* IEEE Trans Neural Netw Learn Syst, 2022. **33**(12): p. 6999-7019.
18. Graves, A.J.S.s.l.w.r.n.n., *Long short-term memory.* Supervised sequence labelling with recurrent neural networks, 2012: p. 37-45.
19. Timmons, P.B. and C.M.J.B.i.b. Hewage, *ENNAVIA is a novel method which employs neural networks for antiviral and anti-coronavirus activity prediction for therapeutic peptides.* Briefings in bioinformatics, 2021. **22**(6).
20. Cao, R., et al., *FFMAVP: a new classifier based on feature fusion and multitask learning for identifying antiviral peptides and their subclasses.* Briefings in Bioinformatics, 2023. **24**(6).
21. Pang, Y., et al., *AVPIden: a new scheme for identification and functional prediction of antiviral peptides based on machine learning approaches.* Briefings in Bioinformatics, 2021. **22**(6).
22. Guan, J., et al., *A two-stage computational framework for identifying antiviral peptides and their functional types based on contrastive learning and multi-feature fusion strategy.* Briefings in Bioinformatics, 2024. **25**(3).
23. Li, Y., et al., *AVP-HNCL: Innovative Contrastive Learning with a Queue-Based Negative Sampling Strategy for Dual-Phase Antiviral Peptide Prediction.* Journal of Chemical Information and Modeling, 2025.
24. Shi, Z. and B.J.a.p.a. Li, *Graph neural networks and attention-based CNN-LSTM for protein classification.* arXiv preprint arXiv, 2022.
25. Vaswani, A., et al., *Attention is all you need.* Advances in neural information processing systems, 2017. **30**.
26. Xu, Y., L. Zhang, and X.J.P.o. Shen, *Multi-modal adaptive gated mechanism for visual question answering.* Plos one, 2023. **18**(6): p. e0287557.
27. Henikoff, S. and J.G. Henikoff, *Amino acid substitution matrices from protein blocks.* Proceedings of the National Academy of Sciences, 1992. **89**(22): p. 10915-10919.
28. Zhang, N., et al., *MutaBind2: Predicting the Impacts of Single and Multiple Mutations on Protein-Protein Interactions.* iScience, 2020. **23**(3): p. 100939.
29. Shrivastava, A., A. Gupta, and R. Girshick. *Training region-based object detectors with online hard example mining.* in *Proceedings of the IEEE conference on computer vision and pattern recognition.* 2016.
30. Hanson, J., et al., *Identifying molecular recognition features in intrinsically disordered regions of proteins by transfer learning.* Bioinformatics, 2020. **36**(4): p. 1107-1113.
31. Xu, Y., et al., *ACVPred: Enhanced prediction of anti-coronavirus peptides by transfer learning combined with data augmentation.* Future Generation Computer Systems, 2024. **160**: p. 305-315.
32. McInnes, L., J. Healy, and J.J.a.p.a. Melville, *Umap: Uniform manifold approximation and projection for dimension reduction.* The Journal of Open Source Software, 2018.
33. Maaten, L.v.d. and G.J.J.o.m.l.r. Hinton, *Visualizing data using t-SNE.* Journal of machine learning research, 2008. **9**(Nov): p. 2579-2605.
34. Selsted, M.E., et al., *Indolicidin, a novel bactericidal tridecapeptide amide from neutrophils.* Journal of Biological Chemistry, 1992. **267**(7): p. 4292-4295.
35. Qureshi, A., et al., *AVPdb: a database of experimentally validated antiviral peptides targeting medically important viruses.* Nucleic acids research, 2014. **42**(D1): p. D1147-D1153.
36. Jhong, J.-H., et al., *dbAMP 2.0: updated resource for antimicrobial peptides with an enhanced scanning method for genomic and proteomic data.* Nucleic acids research, 2022. **50**(D1): p. D460-D470.
37. Kang, X., et al., *DRAMP 2.0, an updated data repository of antimicrobial peptides.* Scientific data, 2019. **6**(1): p. 148.
38. Pirtskhalava, M., et al., *DBAASP v3: database of antimicrobial/cytotoxic activity and structure of peptides as a resource for development of new therapeutics.* Nucleic acids research, 2021. **49**(D1): p. D288-D297.
39. Qureshi, A., N. Thakur, and M.J.P.o. Kumar, *HIPdb: a database of experimentally validated HIV inhibiting peptides.* PloS one, 2013. **8**(1): p. e54908.
40. research, U.C.J.N.a., *UniProt: a hub for protein information.* Nucleic acids research, 2015. **43**(D1): p. D204-D212.
41. Li, W. and A.J.B. Godzik, *Cd-hit: a fast program for clustering and comparing large sets of*

42. Lin, Z., et al., *Evolutionary-scale prediction of atomic-level protein structure with a language model.* Science, 2023. **379**(6637): p. 1123-1130.
43. Bhasin, M. and G.P.J.J.o.B.C. Raghava, *Classification of nuclear receptors based on amino acid composition and dipeptide composition.* Journal of Biological Chemistry, 2004. **279**(22): p. 23262-23266.
44. Ju, Z. and S.-Y.J.G. Wang, *Prediction of lysine formylation sites using the composition of k-spaced amino acid pairs via Chou's 5-steps rule and general pseudo components.* Genomics, 2020. **112**(1): p. 859-866.
45. Cohen, M., V. Potapov, and G.J.P.c.b. Schreiber, *Four distances between pairs of amino acids provide a precise description of their interaction.* PLoS computational biology, 2009. **5**(8): p. e1000470.
46. Chou, K.C.J.P.S., Function, and Bioinformatics, *Prediction of protein cellular attributes using pseudo-amino acid composition.* Proteins: Structure, Function, and Bioinformatics, 2001. **43**(3): p. 246-255.
47. Chou, K.C., *Prediction of protein subcellular locations by incorporating quasi-sequence-order effect.* Biochem Biophys Res Commun, 2000. **278**(2): p. 477-83.
48. Hellberg, S., et al., *Peptide quantitative structure-activity relationships, a multivariate approach.* Journal of medicinal chemistry, 1987. **30**(7): p. 1126-1135.
49. Sandberg, M., et al., *New chemical descriptors relevant for the design of biologically active peptides. A multivariate characterization of 87 amino acids.* Journal of medicinal chemistry 1998. **41**(14): p. 2481-2491.
50. Chen, Z., et al., *iFeature: a Python package and web server for features extraction and selection from protein and peptide sequences.* Bioinformatics, 2018. **34**(14): p. 2499-2502.
51. Saravanan, V. and N.J.M.B. Gautham, *BCIgEPRED—a dual-layer approach for predicting linear IgE epitopes.* Molecular Biology, 2018. **52**(2): p. 285-293.
52. Chen, T., et al. *A simple framework for contrastive learning of visual representations.* in *International conference on machine learning.* 2020. PmLR.
53. Lin, T.-Y., et al. *Focal loss for dense object detection.* in *Proceedings of the IEEE international conference on computer vision.* 2017.
54. Englesson, E. and H.J.a.p.a. Azizpour, *Consistency regularization can improve robustness to label noise.* arXiv preprint arXiv, 2021.
55. Kimura, M. *Understanding test-time augmentation.* in *International Conference on Neural Information Processing.* 2021. Springer.
56. Chicco, D., N. Tötsch, and G.J.B.m. Jurman, *The Matthews correlation coefficient (MCC) is more reliable than balanced accuracy, bookmaker informedness, and markedness in two-class confusion matrix evaluation.* BioData mining, 2021. **14**(1): p. 13.
57. Paszke, A., et al., *Pytorch: An imperative style, high-performance deep learning library.* Advances in neural information processing systems, 2019. **32**.
58. Chen, Z., et al., *iFeatureOmega: an integrative platform for engineering, visualization and analysis of features from molecular sequences, structural and ligand data sets.* Nucleic acids research, 2022. **50**(W1): p. W434-W447.
59. Kingma, D.P., and Jimmy Ba., *Adam: A method for stochastic optimization.* arXiv preprint arXiv, 2014.